\documentclass{article}

\usepackage[final]{nips_2017}

\usepackage[utf8]{inputenc} 
\usepackage[T1]{fontenc}    
\usepackage{hyperref}       
\usepackage{url}            
\usepackage{booktabs}       
\usepackage{amsfonts}       
\usepackage{nicefrac}       
\usepackage{multirow}
\usepackage{microtype}      
\usepackage{float}
\usepackage{graphicx}

\title{PhaseLink: A Deep Learning Approach to Seismic Phase Association}

\author{
  Zachary E. Ross\textsuperscript{1}, Yisong Yue\textsuperscript{2}, Men-Andrin Meier\textsuperscript{1}, Egill Hauksson\textsuperscript{1}, Thomas H. Heaton\textsuperscript{1} \\ \\
  \textsuperscript{1}Seismological Laboratory\\
  California Institute of Technology\\
  Pasadena, CA 91125 \\ \\
  \textsuperscript{2}Department of Computing and Mathematical Sciences\\
  California Institute of Technology\\
  Pasadena, CA 91125 \\
}

\begin{document}

\maketitle

\begin{abstract}

Seismic phase association is a fundamental task in seismology that pertains to linking together phase detections on different sensors that originate from a common earthquake. It is widely employed to detect earthquakes on permanent and temporary seismic networks, and underlies most seismicity catalogs produced around the world. This task can be challenging because the number of sources is unknown, events frequently overlap in time, or can occur simultaneously in different parts of a network. We present PhaseLink, a framework based on recent advances in deep learning for grid-free earthquake phase association. Our approach learns to link phases together that share a common origin, and is trained entirely on tens of millions of synthetic sequences of P- and S-wave arrival times generated using a simple 1D velocity model. Our approach is simple to implement for any tectonic regime, suitable for real-time processing, and can naturally incorporate errors in arrival time picks. Rather than tuning a set of ad hoc hyperparameters to improve performance, PhaseLink can be improved by simply adding examples of problematic cases to the training dataset. We demonstrate the state-of-the-art performance of PhaseLink on a challenging recent sequence from southern California, and synthesized sequences from Japan designed to test the point at which the method fails. For the examined datasets, PhaseLink can precisely associate P- and S-picks to events that are separated by $\sim12$ seconds in origin time. This approach is expected to improve the resolution of seismicity catalogs, add stability to real-time seismic monitoring, and streamline automated processing of large seismic datasets.

\end{abstract}

\section{Introduction}

\begin{figure}[h]
\centering
\includegraphics[width=\textwidth]{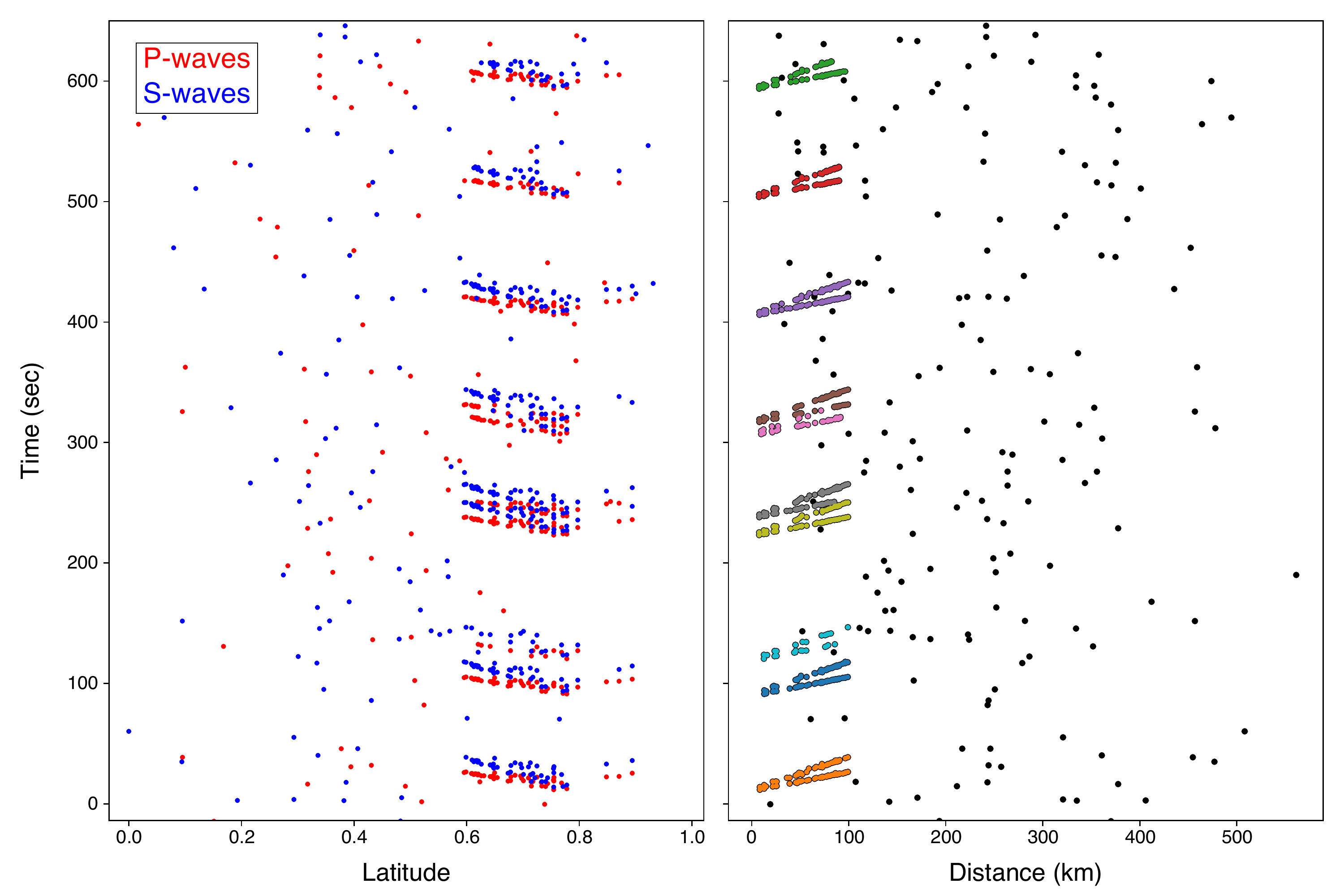}
\label{fig:example}
\caption{Cartoon example of a phase association scenario. Left panel shows the discrete set of picks for the entire network. The number of events is unknown. Right panel shows the output after association and location. Picks colored black are not linked to an event, while colored picks share a common origin.}
\end{figure}

When an earthquake is detected on different stations of a seismic network, it is often desirable to link the observed seismic phases to the earthquake that caused them. Historically, this task was performed by expert seismic analysts who would visually examine the data from different stations, identify seismic phases, and group them together (cf. Figure \ref{fig:example}). As the modern digital era began, seismic networks started to accumulate data in real-time, and it became necessary to develop computer algorithms to automatically process the data.

With the development of the landmark STA/LTA algorithm in seismology \cite{allen_automatic_1978,allen_automatic_1982}, it became possible to detect earthquakes automatically for the first time. This simple method uses the ratio of two moving averages to identify impulsive transient signals and has become the de facto standard for earthquake detection around the world. One major shortcoming of the method is that it will not only identify earthquakes when present, but also any other types of impulsive transient signals that seismometers record. This led to the development of phase association algorithms, which examine combinations of triggers on different stations to see whether any set have arrival time patterns consistent with those of earthquakes \cite{stewart_real-time_1977,johnson_earthworm:_1995,osti_256532,myers_bayesian_2007,reynen_supervised_2017,draelos_new_2015}. The association process therefore evolved from one of simply grouping seismic phases together, to being ultimately responsible for deciding whether an earthquake occurred.

To date, algorithms for phase association all operate using the same fundamental principle. The region of interest is gridded and for each node therein, tentative phase detections within the network are examined to see whether some subset back-projects to a coherent origin. This means that a grid search must be conducted continuously for all new picks that are made. Typically grid associators require extensive tuning of a large number of sensitive hyperparameters, and have numerous ad hoc rules to stabilize potential problems that can arise. Over the years, they have become increasingly sophisticated, with modern variants incorporating Bayesian estimates of pick uncertainties \cite{myers_bayesian_2007}, machine learning  \cite{reynen_supervised_2017}, or multi-scale detection capabilities.

Today, seismologists strive to identify increasingly smaller events that are often at or below the noise level. Resolving this level of detail requires not only increasing phase detection sensitivity, but dealing with the dramatically larger volume of information to be processed in a reliable and rational manner. In particular, since smaller events occur ever more frequently, and therefore are more closely spaced in time, moving forward requires technology that can easily handle the most complicated scenarios encountered at the present.

In recent years, there has been truly astonishing progress within the field of artificial intelligence, most notably in the area of deep learning. Deep learning is a sub-discipline of machine learning that is based on training neural networks to learn generalized representations of extremely large datasets, and has become state of the art in numerous domains of artificial intelligence \cite{lecun_deep_2015}, including natural language processing \cite{sutskever2014sequence}, computer vision \cite{krizhevsky2012imagenet}, and speech recognition \cite{pmlr-v48-amodei16}. It has been recently introduced to seismology, and has already shown considerable promise in performing various tasks including similarity-based earthquake detection and localization \cite{perol_convolutional_2018}, generalized seismic phase detection \cite{ross_generalized_2018}, phase picking \cite{zhu_phasenet:_2018}, first-motion polarity determination \cite{ross_p_2018}, detection of events in laboratory experiments \cite{wu_deepdetect:_2018}, seismic image sharpening \cite{lu_using_2018}, wavefield simulation \cite{moseley_fast_2018}, and predicting aftershock spatial patterns \cite{devries_deep_2018}. 

In this paper, we present PhaseLink, which is a deep learning approach for grid-free earthquake phase association.  Our approach is built upon Recurrent Neural Networks (RNNs), which are designed to learn temporal and contextual relationships in sequential data. We show how to design a training objective that enables the trained RNN to accurately associate phase detections coming from multiple temporally overlapping earthquakes.  Another attractive feature of our approach is that it is trained entirely from synthesized data using simple 1D velocity models.\footnote{ This paradigm is generically known as ``sim-to-real'' in the machine learning community \cite{shotton2011real,shafaei2016play,tobin2017domain,Dosovitskiy17,wang-eisner-2017}. }  Thus, our approach is easily applicable to any tectonic regime by simply training on the synthesized data from the appropriate model, and can also naturally incorporate errors in arrival time picks. The full source code will be publicly available via the Southern California Earthquake Data Center.

\section{Background on Recurrent Neural Networks}

Artificial neural networks are systems that can discover complex non-linear relationships between variables. Fundamentally, they successively transform a set of input values through matrix multiplication and non-linear activation functions into one or more output variables of interest \cite{goodfellow2016deep}. The outputs can be either continuous (regression) or discrete (classification). In supervised learning, the parameters which characterize the non-linear mapping are learned by minimizing the prediction error of the model against the ground truth. The standard type of neural network is today referred to as a fully-connected neural network because each neuron is fully connected to each previous input. Fully-connected networks are excellent at many classification and regression tasks, but have trouble discovering structure in sequential datasets because they lack feedback mechanisms that can enable information to propagate between successive elements of a sequence.

These shortcomings were addressed by the development of the recurrent neural network (RNN) \cite{hopfield_neural_1982}. RNNs allow for information to be passed between successive elements through the use of an internal memory state. This state is dynamically modulated by gates that are themselves composed of neural networks, and control what information is retained along the way. The parameters governing the gates are therefore learned through the training process from the data directly. The outputs of RNNs, which are called hidden states, are very flexible, and could be a single valued output given an input sequence, or a sequence of outputs. To date, RNNs have been applied to variety of settings, including language translation \cite{sutskever2014sequence}, speech synthesis \cite{van2016wavenet}, speech recognition \cite{pmlr-v48-amodei16}, image captioning \cite{you2016image}, and many others. 

The most commonly employed variant of the RNN is the long short-term memory (LSTM) \cite{hochreiter_long_1997} network. These networks have three gates that control the flow of information, and are useful because they are not so susceptible to training issues related to diminishing propagation of information over large sequences. In recent years, another variant called the gated recurrent unit (GRU) \cite {cho_learning_2014} has become popular because it has only two gates instead of three, resulting in fewer parameters and faster training. These types of RNNs are considered state of the art for many problems including speech recognition and language translation. 

Over the years, numerous improvements have been made to these basic types of RNN layers, and one such important development was the bidirectional RNN layer \cite{schuster_bidirectional_1997}. This layer uses two RNNs running in opposite directions so that information from both directions of the sequence is available to make predictions. A common example where this is useful is word prediction, where if a word in the middle of a sentence is missing, it is generally desirable to use the contextual information from the entire sentence to make a prediction, rather than just the words leading up to the missing one.

The outstanding capabilities of RNNs for learning structure in sequential datasets make them a natural choice for the phase association problem, since a set of picks can be viewed as a time-ordered sequence of arrivals. Furthermore, as RNNs process one element of a sequence at a time, they are well-suited for phase association in a real-time seismic network, where phases arrive one at a time.

\section{PhaseLink Framework}
\label{headings}

The PhaseLink approach is designed to solve the phase association problem: given a sequence of \emph{N} picks, determine how many earthquakes (if any) occurred, and which of the \emph{N} picks belong to each respective earthquake. Fundamentally, one can think of phase association as a (supervised) clustering problem of clustering picks to earthquakes that generated them. In contrast to conventional clustering, there is a specific temporal structure to our prediction task, and also the number of clusters is unknown a priori.  For instance, having multiple overlapping earthquakes implies detecting picks coming from different ``clusters''.

Figure \ref{fig:flowchart} depicts the PhaseLink approach, which can be conceptually described in the following steps: 
\begin{figure}[h!]
\centering
\includegraphics[width=\textwidth]{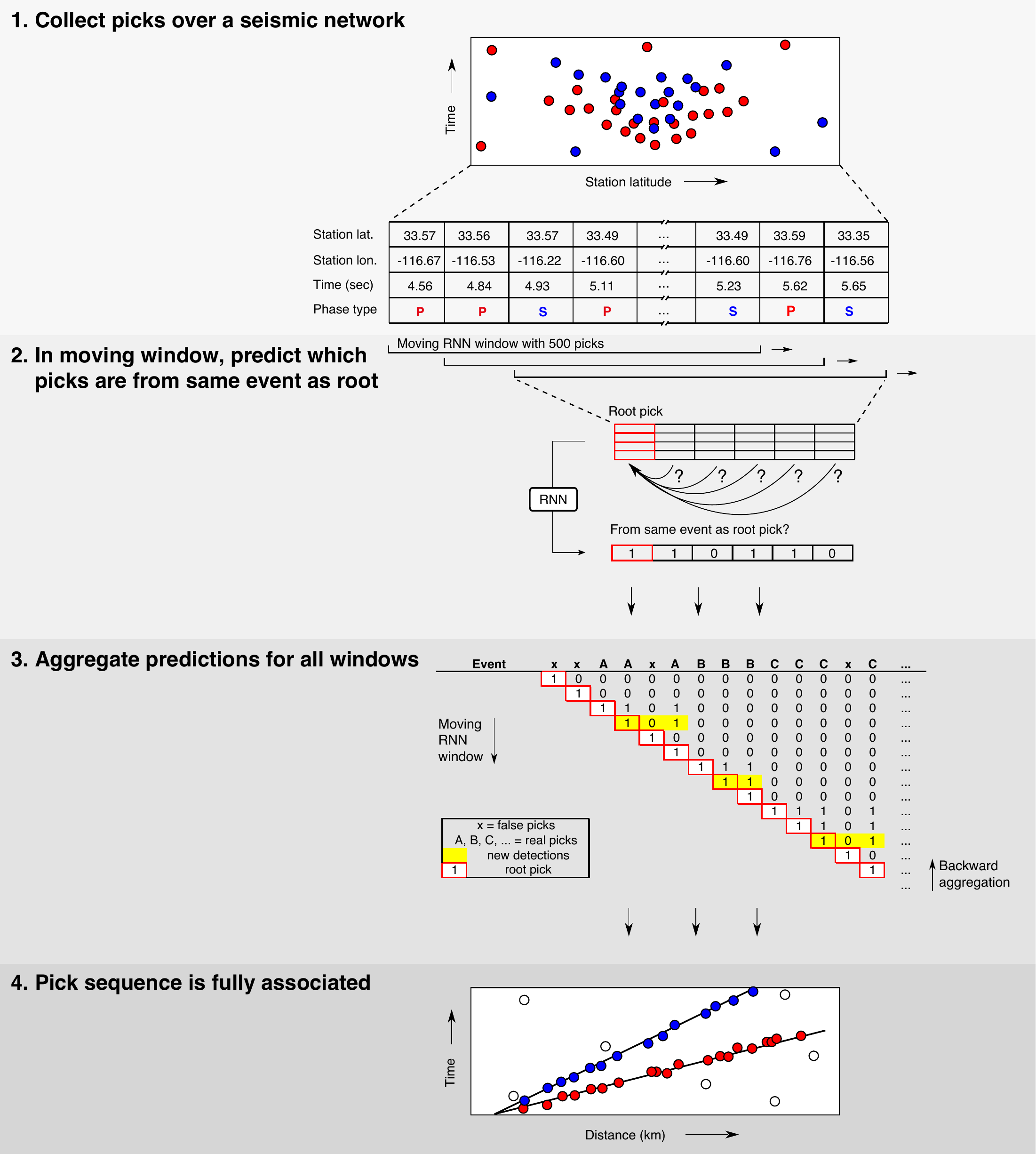}
\caption{Overview of PhaseLink algorithm. A sliding window of picks is iteratively presented to a RNN, which outputs a binary sequence of equal length for each window. These output sequences indicate which picks (if any) are from the same event as the first pick in the window. Each pick in the sequence has five features: latitude, longitude, arrival time, phase type, and a binary padding indicator. The results from all windows are then aggregated to determine distinct clusters of picks (earthquakes detected).}
\label{fig:flowchart}
\end{figure}

\begin{itemize}

\item We are given an input set of picks.  Each pick has as attributes the location (latitude \& longitude) of the station that detected the pick, the time stamp, and phase type (Fig. \ref{fig:flowchart}, step 1).

\item The input pick stream is processed into a sequence of overlapping fixed-length sequential prediction tasks (Fig. \ref{fig:flowchart}, step 2). In particular, the prediction task is whether each pick in the input sequence belongs to the same earthquake that generated the first (root) pick in the sequence, i.e., a sequential binary classification problem. We solve this fixed-length prediction task using RNNs (Section \ref{sec:phaselink_rnn}), and we train the RNNs using synthetic data (Section \ref{sec:phaselink_data}).

\item The overlapping predictions are then aggregated into a single set of pick clusters, where each cluster defines one earthquake (Section \ref{sec:phaselink_aggregate}; Fig. \ref{fig:flowchart}, step 3).

\end{itemize}

By decomposing the problem in this way, PhaseLink can, in principle, handle any number of overlapping clusters. Conceptually, the reduced prediction task  is based around a reference point and classifies a temporal neighborhood of points as belonging to the same cluster as the reference point.  A somewhat similar idea was proposed in supervised clustering approaches that utilize must-link and cannot-link constraints
\cite{wagstaff2001constrained,bilenko2004integrating}, although those approaches are more geared towards learning a metric space rather than directly solving the clustering problem. Furthermore, our PhaseLink approach can exploit a natural temporal locality structure to further constrain the prediction task. Another benefit of directly considering the co-clustering prediction problem is that we can tolerate false picks (those that do not belong to any cluster/earthquake). A final benefit of this decomposition is that PhaseLink can utilize off-the-shelf RNN implementations, which leads to significantly reduced system engineering overhead.

\subsection{RNN Architecture}
\label{sec:phaselink_rnn}
We designed a deep RNN consisting of stacked bidirectional GRU layers (Table \ref{table:mod_arch}). The network takes as input fixed-length sequences of picks, and outputs a sequence of identical length (Fig. \ref{fig:flowchart}, step 2). The output sequence is binary valued, with a value of 1 indicating that a given pick belongs to the same event as the root pick (the first pick of the sequence, $Y_0$), and a value of 0 indicating that the two picks are unrelated. A sigmoid activation function is applied to the final output at each time step to squash the value into the range $[0, 1]$. 

We apply the network to a sliding window of picks by incrementing over the entire sequence, shifting the window by one pick at a time. For the remainder of this paper, we refer to a fixed length sliding window of $n_p$ picks as a \textit{sub-sequence}. Here we use $n_p$ = 500. After predictions have been made for a sub-sequence we drop the root pick and take the next pick as the root for the new sub-sequence. For each root pick we obtain a set of binary predictions about which of the following picks in the sub-sequence are related to the root. 

\begin{table}[h]
\centering
\caption{Model architecture}
\begin{tabular}{ | c | c | c | }
   \hline
   Layer type & Units & Activation function\\ \hline
   Bidirectional GRU & 200 & sigmoid/tanh \\ \hline
   Bidirectional GRU & 200 & sigmoid/tanh \\ \hline
   Dense & 1 & sigmoid \\
   \hline
\end{tabular}
\label{table:mod_arch}
\end{table}

Each of the picks in a sub-sequence is characterized by five input features, resulting in an input feature set with dimensions ($n_p$, 5). The first two features are the latitude and longitude coordinates of the station that the pick was made on, which are both normalized to be in the range [0, 1] such that the range spans the full dimensions of the seismic network. The third feature is the time of the pick, which is defined relative to the root pick within the sub-sequence. Here, we normalize the time values by a pre-defined maximum allowed value for picks to be included in a sub-sequence, which is chosen to be 120 seconds. This value is somewhat arbitrarily chosen, but ends up being not too important. The normalization ensures that this feature does not bias the training process. We discard any picks within the sub-sequence that are larger than 120 s, and pad the remainder of the feature window with zeros. The value of 500 picks is chosen loosely to correspond to the maximum number of picks that we expect to have within any 120 s window, which could vary depending on the problem. The penultimate feature is a binary value indicating the phase type, where a value of 0 means a P-wave, and a value of 1 means an S-wave. Lastly, we have another binary indicator variable for whether a given pick is a zero-padded placeholder.

\subsection{Aggregating Predictions}
\label{sec:phaselink_aggregate}
We now describe the final stage of PhaseLink, where the link predictions from each sub-sequence are aggregated to formally detect earthquakes. The output of the RNN is a prediction matrix that describes the link between each pick of a sub-sequence and its root pick (Figure \ref{fig:flowchart}, step 3). In order to assign picks to individual events, rather than to sub-sequence root picks, we cluster linked picks by incrementing backwards over the prediction matrix. This is performed as follows:

\begin{itemize}

\item For each sub-sequence, a cluster nucleates if at least $n_{nuc}$ picks have predicted labels of 1. Only those picks labeled 1 are retained in the cluster, and all others are discarded.

\item Once a cluster has nucleated, the set intersection is separately determined between it and every existing cluster of picks that arose from other sub-sequences.

\item The existing cluster with the most picks in common is identified, and if this number is greater than $n_{merge}$, the two clusters are merged.

\item After performing these steps for all sub-sequences, each remaining cluster is retained if the cluster size is at least $n_{min}$ picks.

\end{itemize}

In this paper, we use $n_{nuc}$ = 8, which was chosen to maximize the detection performance for the datasets used herein; varying this hyperparameter in the range 4-8 leads to relatively little change in performance on these datasets. Since the root pick is always linked to itself, the largest possible value of $n_{merge}=n_{nuc}-1$. Here, we use this maximum value, $n_{merge}=7$. Since merging clusters is performed by identifying the existing cluster with the most picks in common, there is the possibility that a pick could end up in two separate clusters; however in our testing, this is extremely uncommon. $n_{min}$ is the most sensitive of the three hyperparameters, and its effect on the performance is examined in detail in the next section.

After applying the aforementioned steps, the PhaseLink algorithm is completed and the sequence is fully associated. We note that no hypocenters have been determined during this process for any events, whereas other associators jointly solve for a location as part of the detection process, which is a more challenging problem. Thus far, we have only discussed the method and how it is to be used; in the next section, we describe a scheme for generating the training data for the RNN.

\subsection{Sim-to-Real Training}
\label{sec:phaselink_data}
The RNN described in Section \ref{sec:phaselink_rnn} could, in principle, be trained with real seismic phase data. However, even in the most seismically active regions, the available data may barely be enough to effectively train such a deep network. Since the network only requires phase arrival times and station geometries we can instead generate synthetic training datasets of arbitrary size. The key intuition here is that the supervised clustering problem solved by PhaseLink need not require fully realistic earthquake data to train an accurate predictor.

\subsubsection{Synthetic Data Generation}
We develop a simple scheme for generating large datasets of synthetic pick sub-sequences using a 1D layered model. The goal is for the neural network to learn the essential physics of wave propagation from the synthetic data, so that this knowledge can be directly applied to real data. To do so we define a set of rules from which random pick sub-sequences are generated. We use uniform distributions for all random quantities and denote this distribution as $U$. The rules to generate a single sub-sequence realization are as follows:

\begin{enumerate}

\item The initial number of events is chosen from $U \sim [0, 20]$.

\item A random hypocenter is initially assigned to all events. The latitude and longitude are each drawn from $U \sim [0, 1]$, while the depth is drawn from $U \sim [0, 25]$ km.

\item At a probability of 10\%, each event is then separately reassigned a new hypocenter to produce events that overlap in time, but originate in different parts of the network.

\item The first event is assigned an origin time from $U \sim [-60, 60]$ s, enabling the possibility of the event to have an origin time before the sub-sequence starts.

\item The origin times for all subsequent events are chosen such that the time between consecutive events is $U \sim [3, 20]$ s

\item The maximum source-reciever distance for each event is drawn from $U \sim [20, 100]$ km.

\item Arrival times are calculated with a 1D model for all source-receiver combinations within the chosen maximum distance.

\item Picks are randomly discarded with $\textnormal{Pr}=0.5$ to add variability to the station distribution.

\item Arrival time errors are added to each pick and drawn from $U \sim [-0.5, 0.5]$ s.

\item A random number of false picks drawn from $U \sim [0, 500]$ are randomly distributed across the network, with origin times drawn from $U \sim [0, 120]$ s.

\item Picks outside of the time window of $[0, 120]$ s are removed.

\item All remaining picks are sorted in time and the first 500 are retained. If fewer than 500 exist, the feature matrix is padded accordingly.

\end{enumerate}

The ability to generate synthetic training data for the model to learn from has several important advantages. First, since a simple 1D layered velocity model is used to calculate arrival times, the method is easily applied to most tectonic regimes with relatively little knowledge required about the velocity structure. Second, errors can be directly added to the synthetic arrivals such that the model learns to deal with them in a rational way when examining real data. Third, an unlimited amount of training data can be generated, which can prevent overfitting and regularization issues during the training process.

\begin{figure}[t]
\centering
\includegraphics[width=0.75\textwidth]{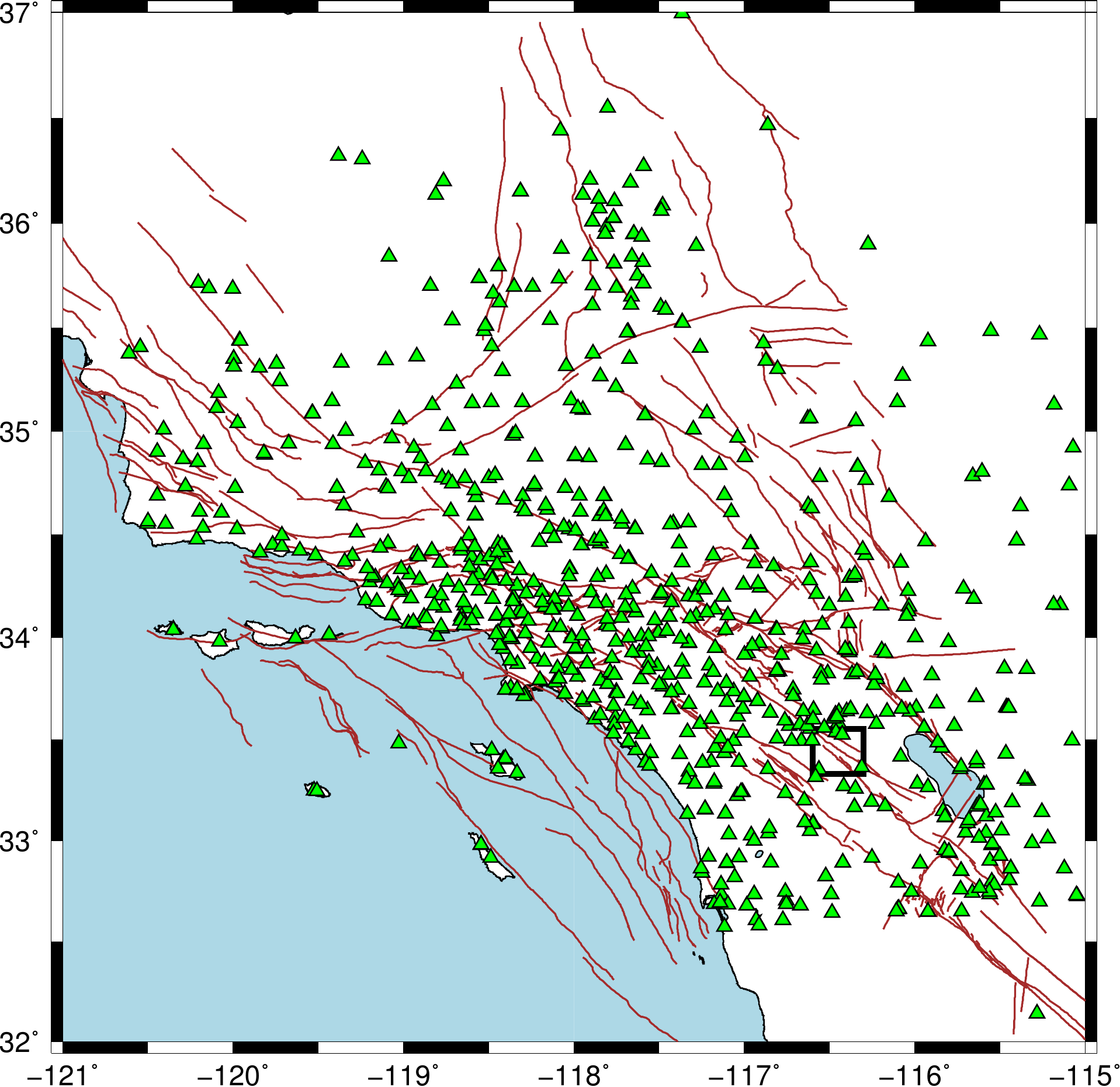}
\caption{Map of southern California and the SCSN station distribution. The entire region shown defines the boundaries for generating training data. The black box indicates the region from which events were selected for the Borrego Springs tests.}
\label{fig:socal_map}
\end{figure}

\begin{figure}[h]
\centering
\includegraphics[width=0.75\textwidth]{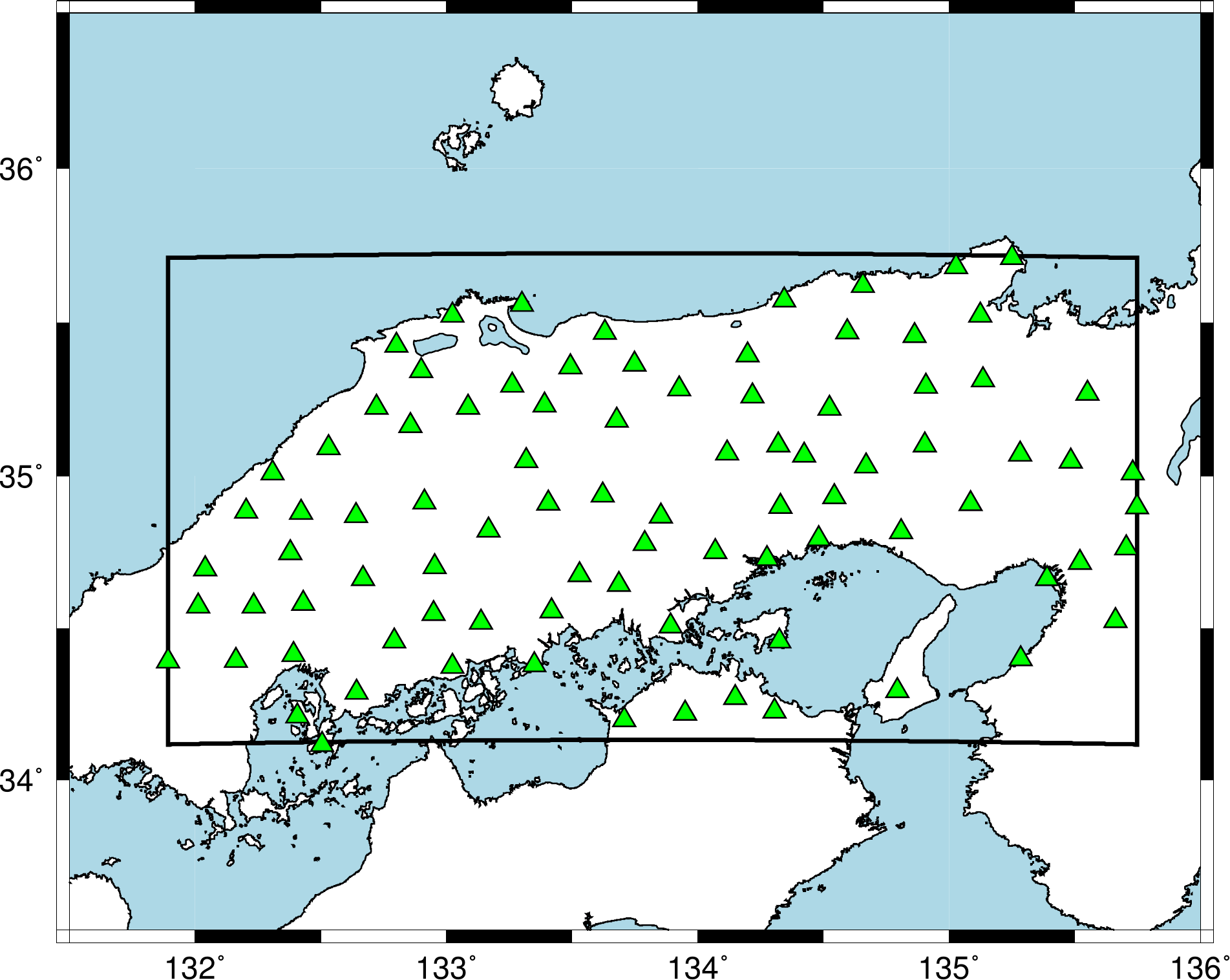}
\caption{Map of southwestern Japan and station distribution (green triangles). The region used for generating synthetic events is indicated by the solid black line.}
\label{fig:japan_map}
\end{figure}

We produce two separate training datasets in this study. The first is for southern California, using the exact station distribution from the 811 past and present stations of the SCSN (Figure \ref{fig:socal_map}). We use a simple 1D velocity model for southern California \cite{hadley_seismic_1977}. The second is for southwestern Japan, using 88 stations of the Hi-net seismic network (Figure \ref{fig:japan_map}). For the Japanese dataset, we use a 1D model for southwest Japan \cite{shibutani_high_2005}.

\begin{figure}[t]
\centering
\includegraphics[width=0.85\textwidth]{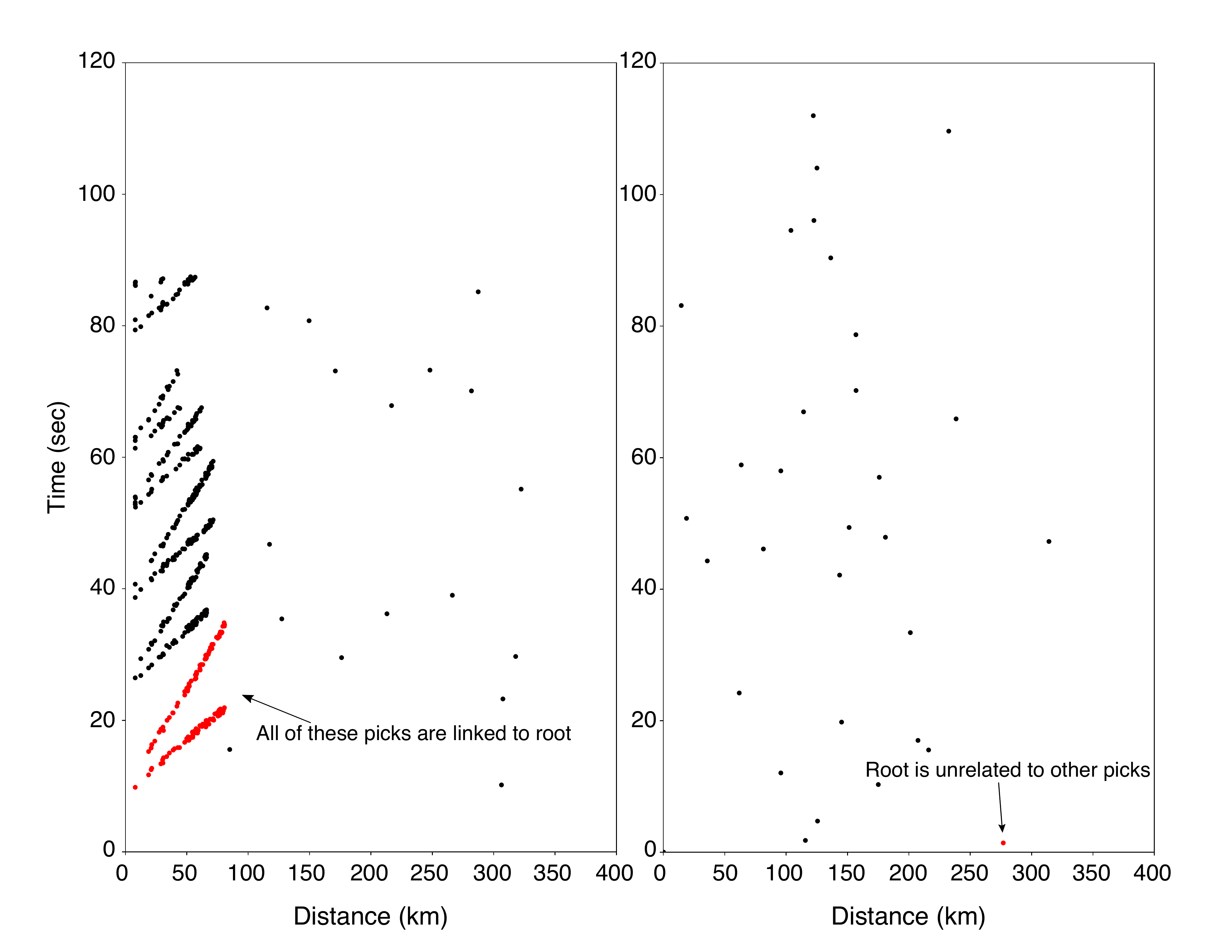}
\caption{Examples of synthetic training sub-sequences. Red picks are linked to the root, while black picks are unrelated to the root.}
\label{fig:synth}
\end{figure}

Examples of two sub-sequence realizations are shown in Figure \ref{fig:synth}. The labels of each sub-sequence, $Y_i$, depend on whether the first pick, $Y_0$, is associated to an earthquake or not. If it is, then it and all picks associated with the event are given a label of 1, while all other picks are given a label of 0. Otherwise, $Y_0$ is the only non-zero value. We then repeat all of these steps 12 million times to generate a total of (up to) 6 billion picks.

\subsubsection{Training the RNN}
\begin{figure}[t]
\centering
\includegraphics[width=0.75\textwidth]{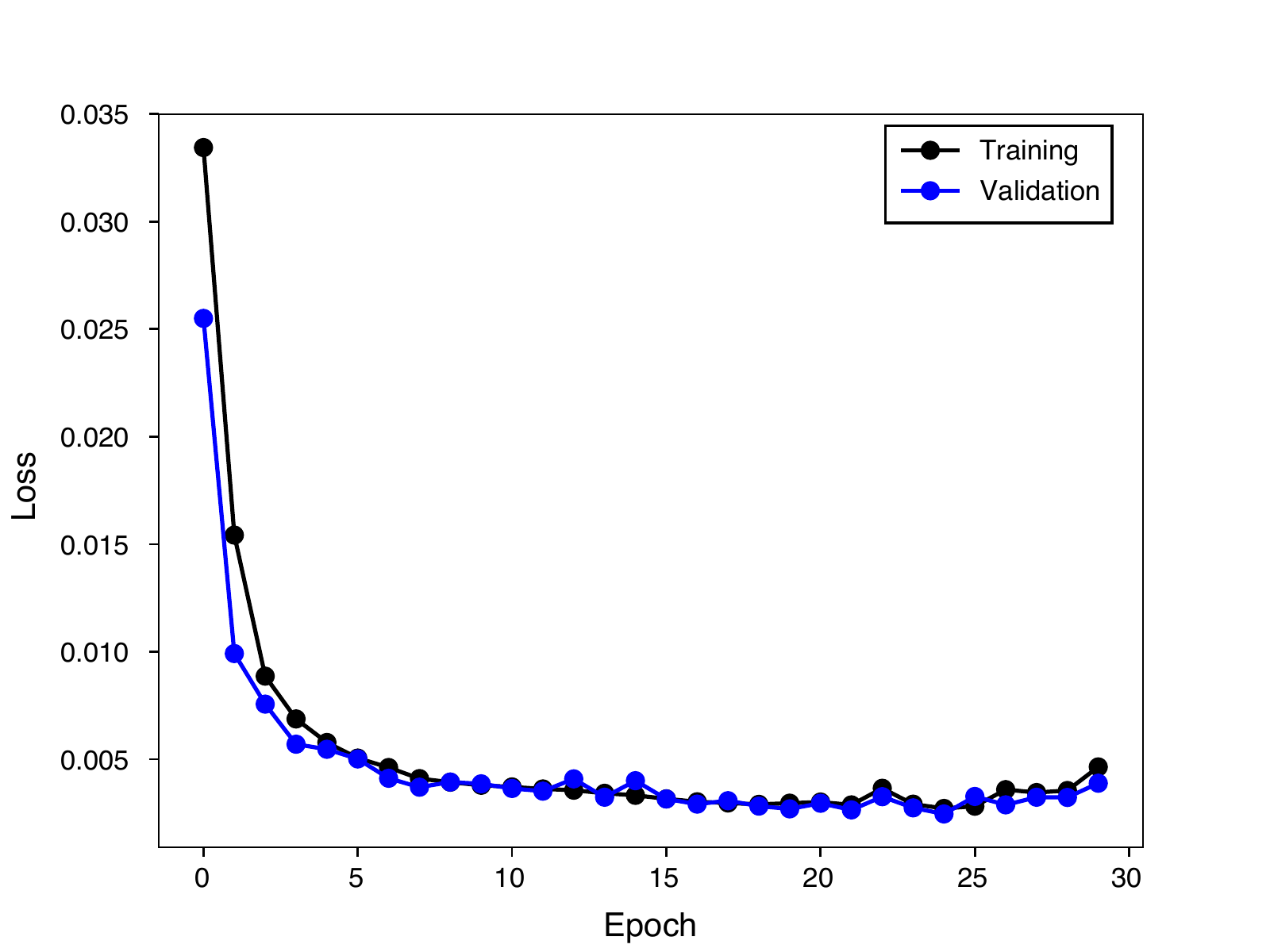}
\caption{Training and validation loss histories.}
\label{fig:losses}
\end{figure}

Given the generated datasets, we can train the RNN using any off-the-shelf machine learning package.  The 12 million sub-sequences are designed to represent a wide variety of possible phase arrival time scenarios from all over southern California and Japan. From here, we can train the RNN to link phases together. We randomly split our 12 million sub-sequences into training (75\%) and validation (25\%) sets. To train the model, we use the binary cross-entropy loss function,
\begin{equation}
L = -\frac{1}{N} \sum_{i=1}^{N} \big[y_i\:\textnormal{log}(p_i) + (1 - y_i)\:\textnormal{log}(1-p_i)\big],
\end{equation}
where $N$ is the number of samples, $y_i$ is the true label of the $i$th sample, and $p_i$ is the predicted probability for the $i$th sample. The model was trained using three NVIDIA GTX 1060 GPUs, the Adam optimization algorithm \cite{kingma_adam:_2014}, and a mini-batch size of 96. On the synthetic validation data, the model achieves a categorical accuracy of 99.92\%. The loss histories for the training and validation datasets are shown as a function of the number of training epochs (cycles through the entire dataset) in Figure \ref{fig:losses}.

\section{Results}
\label{sec:phaselink_results}
In this section, we examine the performance of PhaseLink under a variety of scenarios. All of the tests are conducted in a controlled manner, such that ground truth is known for every single pick. This enables a detailed assessment of the performance at the individual phase level for real sequences of picks, as well as sequences designed to test the point at which the method breaks down. It furthermore allows for a rigorous direct comparison of PhaseLink with existing grid association methods.

\subsection{Application to 2016 Borrego Springs sequence}
We apply PhaseLink to the 2016 Borrego Springs sequence, which occurred in the San Jacinto fault zone in southern California \cite{ross_abundant_2017}. During the period 2016-06-01 to 2016-06-31, 1708 earthquakes were identified by the Southern California Seismic Network (SCSN) within the study area (Figure 3, black box), and 73,353 phases were picked by seismic analysts. We use all of these events and phases in this study. The phase data are publicly available from the Southern California Earthquake Data Center. For a controlled testing environment, we reconstruct the exact sequence of 73,353 picks for all events. Then we add in an equivalent number of false picks (73,353) uniformly distributed over the seismic network in time, i.e., picks that do not belong to an earthquake. This results in 50\% of the picks in the sequence being false, but the effect is not uniform over time since the number of events (and therefore the number of picks) decreases with time after the mainshock. This has the overall effect of mimicking a real seismic network, where the system is dominated by real picks during a swarm, and later dominated by false picks the rest of the time.

First, we examine the performance of the neural network alone, without the clustering step included (Table \ref{table:perf}). Precision is defined as the ratio of true positives to the true positives plus false positives. Recall is defined as the ratio of true positives to the true positives plus false negatives. The high precision for false picks (i.e. with true label = 0) shows that it is relatively rare for the network to assign a false pick label to real picks. Similarly, the high recall suggests that false picks are rarely assigned label 1. For the real picks (true label = 1), the high precision implies that the network rarely incorporates false picks into sequences of real picks. The lower recall, on the other hand, implies that it quite often discards real picks as false. Since the number of unrelated picks is much larger than that of related picks, these mis-classifications affect the real pick recall much more than the false pick precision.     

However, this performance is only considering association relative to the root picks of individual sub-sequences. As we will demonstrate below the clustering scheme described in section 3 successfully recovers many of these missed picks, since a pick only needs to be associated correctly in one of the sub-sequences it appears in.

\begin{figure}[t]
\centering
\includegraphics[width=\textwidth]{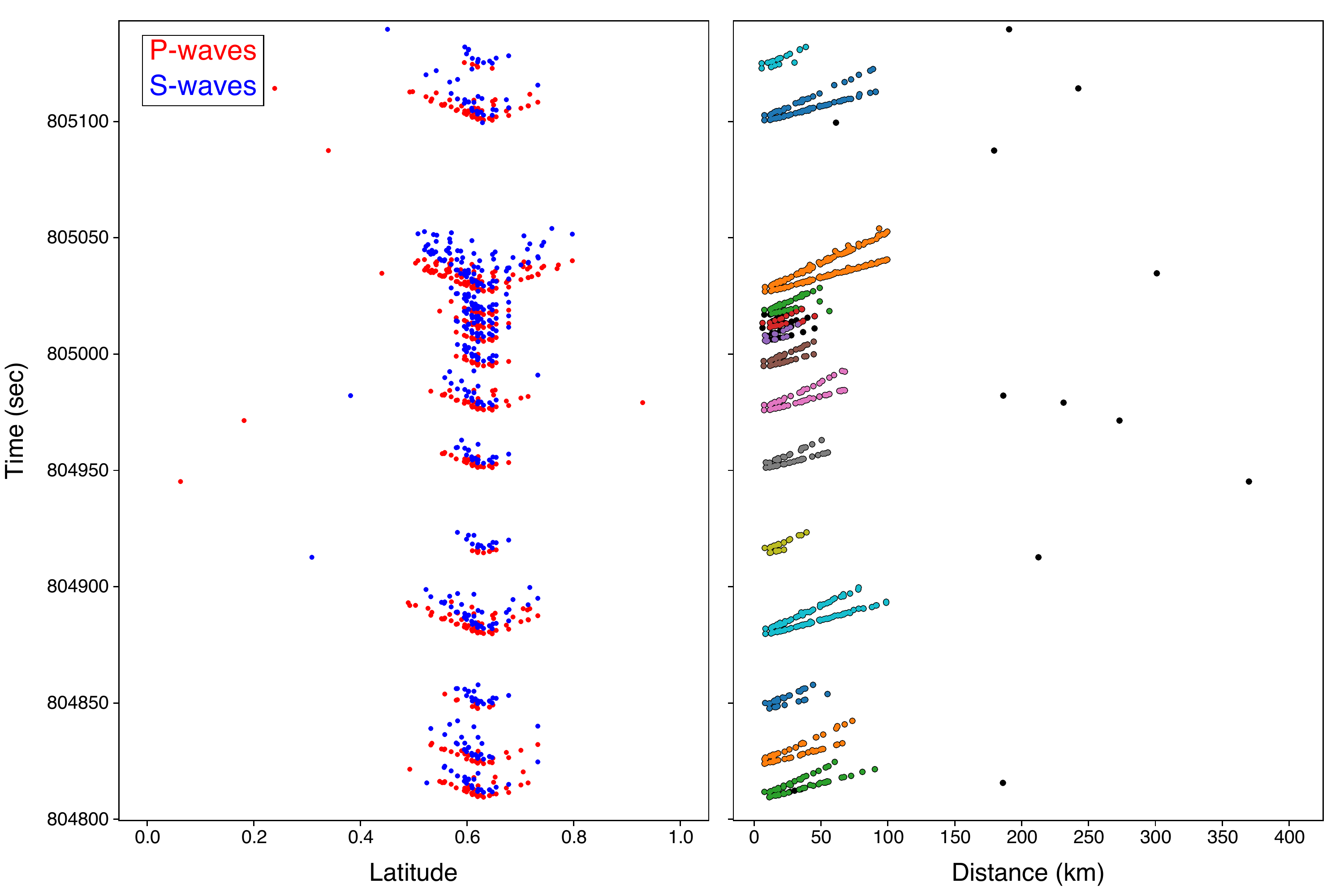}
\caption{Sample of events detected during the 2016 Borrego Springs sequence. Within a span of 350 seconds, there are 14 events detected successfully. Left panel shows picks across the entire network for a given window of time. Latitude is normalized to be in the range (0, 1). Right panel shows output events detected and respective phases associated to each event (colored circles). Black circles indicate phases that are left unassociated by the algorithm.}
\label{fig:borrego_ex}
\end{figure}

Next, we examine the performance on a challenging segment of the dataset in Figure \ref{fig:borrego_ex}, where there are 14 earthquakes within a time span of roughly six minutes. In the right-hand panel, phases that are associated to the same event have circles with the same color. It can be seen that the method does an excellent job, even down to the point when the events are spaced only 5-10 seconds apart in origin time.

\begin{table}[tb]
\centering
\caption{RNN performance on validation dataset (individual phases)}
\begin{tabular}{ | c | c | c | c | }
   \hline
   True label & Precision & Recall & \# samples \\ \hline
   0 & 0.99 & 0.99 & 71116302\\ \hline
   1 & 0.98 & 0.96 & 2236698\\
   \hline
\end{tabular}
\label{table:perf}
\end{table}

\begin{figure}[h]
\centering
\includegraphics[width=0.7\textwidth]{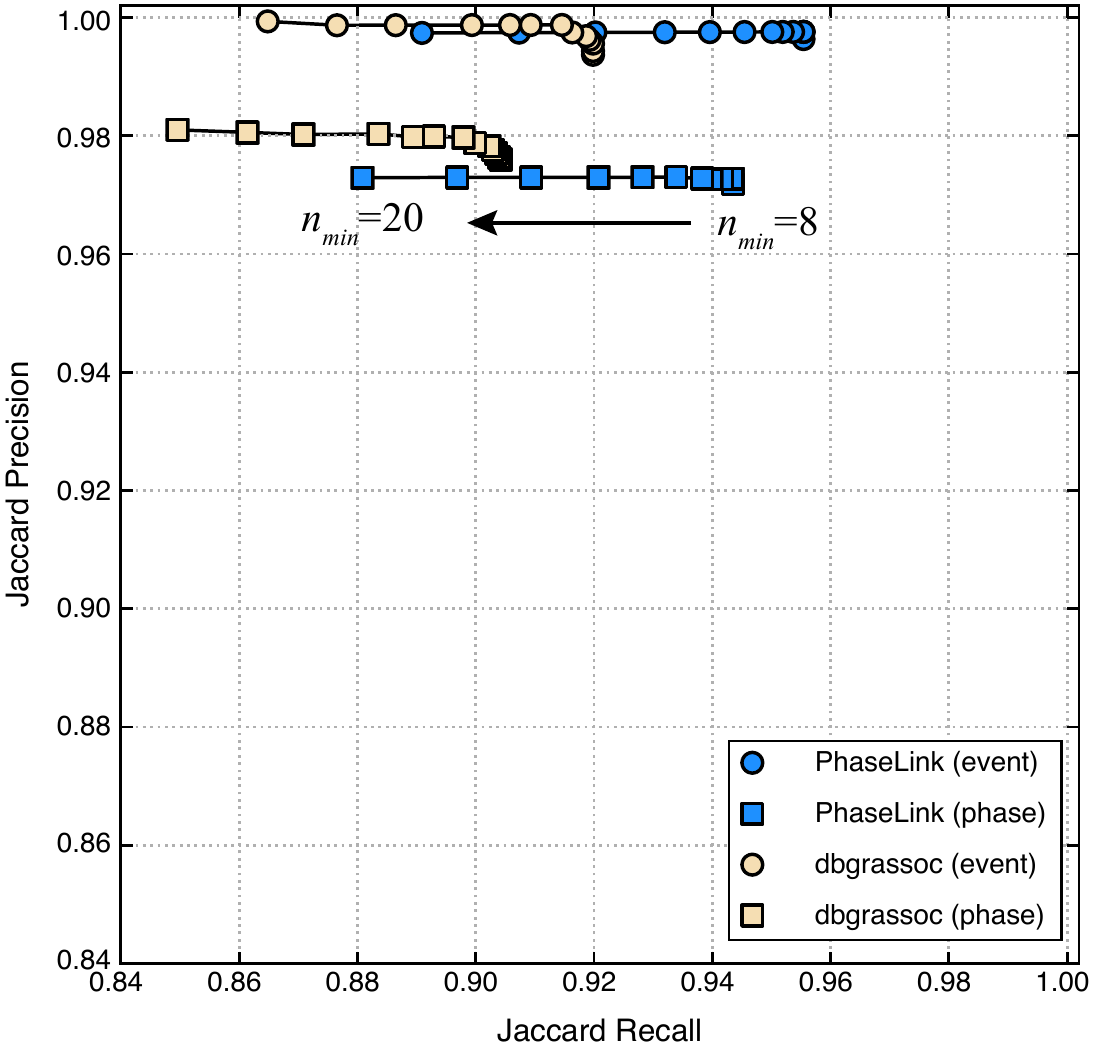}
\caption{Precision-recall tradeoff curve for event detection and phase association performance on the 2016 Borrego Springs sequence. Each point represents a value of $n_{min}$ from 8-20.}
\label{fig:borrego_perf}
\end{figure}

Figure \ref{fig:borrego_perf} demonstrates the outstanding performance of the complete algorithm in the context of detecting earthquakes, rather than individual phases. This precision-recall curve illustrates the inherent trade-off when the minimum number of picks per cluster, $n_{min}$, is varied. To determine whether the $k$th cluster of picks, $A_k$, corresponds to a successful event detection, we define the Jaccard precision between it and all clusters of picks in the ground truth, $B_i$,
\begin{equation}
J_k^p = \max_{i=1}^{c}\frac{A_k \cap B_i}{A_k \cup B_i},
\end{equation}
where, $c$ is the total number of events in the ground truth. We also define the Jaccard recall,
\begin{equation}
J_i^r = \max_{k=1}^{d}\frac{A_k \cap B_i}{A_k \cup B_i},
\end{equation}
where $d$ is the number of detected events. If $J_k^p \ge 0.5$, we consider the detection successful. This means that at least 50\% of the picks in the predicted cluster are common with a single event in the ground truth.  

For all values of $n_{min}$, the precision is $>0.996$. However, raising $n_{min}$ decreases the recall from 0.956 to ultimately 0.891, because real clusters that have fewer than $n_{min}$ picks get discarded. High values of $n_{min}$ decrease the algorithm's ability to detect and associate weakly recorded small earthquakes for which only small numbers of phase detections are available. However, at least for this test sequence, it appears that $n_{min}\sim8$ is sufficient to get excellent association performance from the algorithm.

These performance numbers are in terms of event declarations, but it is also possible to evaluate the performance of phase associations as well. To do this, we average $J^p$ over the $d$ detected events, and average $J^r$ over the $c$ ground truth events. Together, these average quantities represent the precision and recall at an individual phase level, rather than an event level. These values are also shown in Figure \ref{fig:borrego_perf} for the same range of $n_{min}$ values, indicating that not only is the method detecting events well, but that it also reliably associates phases.

In developing a new method, it is also important to benchmark its performance against that of existing methods. Here, we compare the performance of PhaseLink against the grid associator, \textit{dbgrassoc}, from the Antelope Environmental Monitoring Software package (BRTT Inc.). \textit{dbgrassoc} is currently used by real-time seismic networks around the world, as well as researchers working with previously collected datasets in an offline mode. The program uses a pre-defined travel time grid that is set up over the region in which earthquakes are to be detected. There are a number of sensitive hyperparameters that control the detection process including the minimum number of picks, whether S-waves are to be included and how to deal with them, travel time residual limits, and the clustering time window. Once an event is detected, it also re-examines previous detections to see if the new event should be merged with another, or extra phases can be added in. For this comparison, we use the exact settings employed by a detection study in the San Jacinto fault zone \cite{ross_improved_2016}, which are very similar to those used internally for real-time operation by the Anza seismic network. This ensures that \textit{dbgrassoc} is correctly calibrated and that the comparison is fair.

We applied \textit{dbgrassoc} to the same sequence of picks as used with PhaseLink, and the results are shown in Figure \ref{fig:borrego_perf}. At an event level, \textit{dbgrassoc} and PhaseLink have nearly identical precision ($>0.996$), but the recall for PhaseLink is significantly higher (0.956 vs 0.919). When considering phase association performance, rather than event detection performance, \textit{dbgrassoc} has slightly higher precision (0.976 vs 0.9718). However, PhaseLink has a much higher recall of 0.955, whereas \textit{dbgrassoc} has a value of 0.904. Together, these tests show that PhaseLink detects significantly more events (63) in the sequence and correctly associates more phases to each event (7,482), without sacrificing precision.

It should be noted that while this is a challenging sequence, the events were all large enough for humans to manually pick them. As seismologists seek to detect increasingly smaller magnitude events, not only will humans be unable to pick many of them, but the volume of data may be an order of magnitude larger \cite{shelly_fluid-faulting_2016,ross_generalized_2018}, with the timing between events significantly shorter. It is here that the full potential of PhaseLink is realized as compared with grid associators.

\subsection{Stress testing PhaseLink in Japan}
We next run PhaseLink through a series of tests to examine the conditions under which the algorithm finally breaks down. To do this, we construct random sequences of earthquakes synthetically with progressively more difficult scenarios: in each test, the average time between events is shortened relative to the previous test. We use a subset (88) of the Hi-Net seismic network stations from Japan that are concentrated around the Chugoku region (Figure \ref{fig:japan_map}). We generate 8 random sequences of 5000 earthquakes each within the region. For each sequence, the time between consecutive events is randomly drawn from a uniform distribution, with a fixed minimum value of 0 seconds, and a maximum value of 10, 12, 16, 20, 24, 32, 64, and 128 seconds, respectively. We then define $\overline{\Delta t_o}$ as the average time between events over the 5000 events in the sequence. Thus, in the hardest sequence, $\overline{\Delta t_o}=5$ s, and in the easiest sequence, $\overline{\Delta t_o}=64$ s. Each event is given a random hypocenter within the box defined by Figure \ref{fig:japan_map}, and the maximum source-receiver distance is drawn from $U \sim [20, 100]$ km.  Arrival times are calculated with a 1D model for the region \cite{shibutani_high_2005}, and there are approximately 100,000 picks in each of the sequences. Since the minimum time between events is 0 for all sequences, this allows for the possibility that some events are simultaneous in time, but with different hypocenters, which is an extremely challenging problem. We then add random errors to the picks from $U \sim [-0.5, 0.5]$ s. Then, we run PhaseLink on the 8 sequences and calculate the performance against the ground truth for each one.

\begin{figure}[h]
\centering
\includegraphics[width=0.8\textwidth]{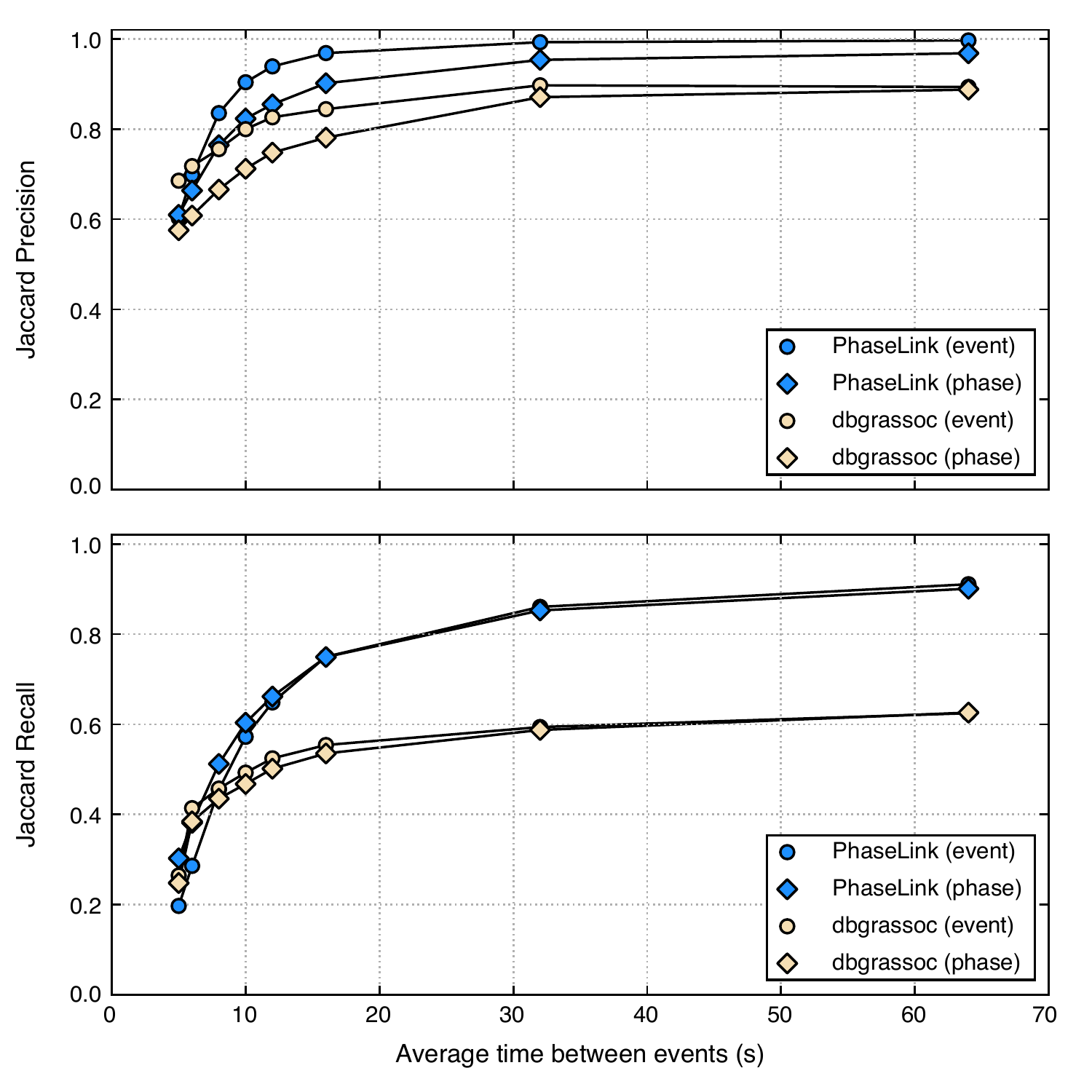}
\caption{Stress testing PhaseLink and dbgrassoc on synthetic earthquake sequences in Japan. As the average time between events in a sequence decreases, the precision and recall decrease for both methods. PhaseLink significantly outperforms dbgrassoc even on the easiest cases in the test dataset. These tests indicate that for this dataset, PhaseLink begins to break down when the average time between events approaches 10-12 seconds.}
\label{fig:japan_perf}
\end{figure}

Figure \ref{fig:japan_perf} shows the results of the stress test on the synthesized Japanese data. For the easiest test, when $\overline{\Delta t_o}=64$ s, the event precision and recall are 0.996 and 0.911 respectively. For the most difficult test, when events are 4 seconds apart on average, the performance is very poor, with the precision dropping to 0.603, and recall having a value of only 0.197. The figure indicates that the method stops achieving reasonable performance when the events are approximately 10-12 seconds apart, although this is somewhat subjective as there is not a steep falloff point in the curves. Regardless, a sequence of 5000 events that are 10 seconds apart on average is a very extreme scenario, and demonstrates the impressive performance of PhaseLink as an association algorithm. We also performed the same tests with \textit{dbgrassoc}, and the performance is shown in Figure \ref{fig:japan_perf} as well. It is clear that even for the easiest test that \textit{dbgrassoc} missed nearly 40\% of the events (compared with 9.5\% for PhaseLink) and had a false positive detection rate that is nearly 10 times higher than PhaseLink. This is true at a phase association performance level as well. We conclude that PhaseLink significantly outperforms \textit{dbgrassoc} when events are densely clustered in time.

\section{Discussion}
Earthquake phase association is an essential task in seismology that has been around for decades but still remains a challenge at the present day, most noticeably during dense earthquake sequences. PhaseLink is distinctly different from previous methods for phase association because it does not need to search over a hypocenter grid for a solution. Instead, it uses the capabilities of deep neural networks to learn the patterns that seismic waves make as they propagate across a seismic network. The results indicate that the method has all of the key functionality necessary to solve this difficult problem in an efficient and robust way, and that recurrent neural networks are capable of learning the basic physics of seismic wave propagation in the Earth. Here, we discuss several important aspects about the algorithm in more detail.

\subsection{Potential future modifications} It is reasonable to expect that some modifications to the methodology will be necessary in the future. What distinguishes PhaseLink from other association algorithms is that when persistent problems are identified, examples of such cases can be added to the training dataset so that the neural network learns how to deal with them. This is a fundamentally different approach from grid-based association algorithms, where addressing problematic cases must be done by either tuning hyperparameters or modifying the algorithm altogether.

We have demonstrated the usage of PhaseLink only in the context of a local seismic network. Applying the method to regional or teleseismic scales may require additional modifications, but we do not see anything that would preclude the method from being used at these scales. At such distances, there are often more phase types available that could be explicitly incorporated into the problem. Since PhaseLink automatically associates phase picks together without solving for a location, once a hypocenter is determined, it is possible to then perform a quick back-projection of all nearby phases to clean up or add in any split events or unassociated phases.

Traditionally, associators have determined the type of phase by checking different combinations of moveout-trigger patterns and identifying which phase is the most likely. Due to recent advances in phase identification with deep learning \cite{ross_generalized_2018,zhu_phasenet:_2018}, it is now possible to determine the phase type as part of the detection process, rather than leave this as something to be determined by an associator. We have designed PhaseLink to therefore take labeled phases as input, with the expectation that the labels result from one of these types of algorithms. When combined with these other deep learning approaches, it is possible to have an end-to-end detection pipeline.

\subsection{Comparison of PhaseLink and Grid Associators} Grid associators link phases together while simultaneously locating an event. As this is essentially solving two (interdependent) problems, it can lead to false event detections if a random subset of phases have a moveout that is consistent with a single grid node acting as a hypocenter. Furthermore, the large number of ad hoc hyperparameters that are necessary to stabilize the association scheme can lead to various types of errors. PhaseLink does not use a grid or locate earthquakes; rather, it solves a simpler problem: group phases together with arrival time patterns that are generally consistent with seismic wave propagation, irrespective of the hypocenter. This will enable robust phase association, and these phases can then be presented to a location algorithm to determine the best possible hypocenter.

\subsection{Advantages of training with synthetic data} A key advantage of the algorithm is that it can be trained using only synthetic data. Because of this it is directly applicable to regions for which there is insufficient real phase data. This includes small networks as well as temporary ones that have never recorded data before. The only requirement is a sufficiently accurate velocity model for generating the training data, which for most cases, is probably a 1D layered model.

Another distinct advantage of PhaseLink is that picking errors can be explicitly accounted for in the training data. Here, we chose a rather extreme case by drawing errors from $U \sim [-0.5, 0.5]$ s to force the neural network to learn how to deal with this complexity. This also has the advantage of learning to rationally deal with genuine perturbations in arrival times due to 3D structure. We note that in a limited sense, this attribute is similar to the BayesLoc algorithm \cite{myers_bayesian_2007}, which uses a Bayesian formalism to account for the possibility of errors in phase picks and other factors for association purposes.

The developed algorithm has the potential to significantly improve automated processing of large seismic datasets. This is important both for researchers who are looking to build catalogs from long-existing or newly collected datasets, as well as seismic networks that are responsible for routine seismic monitoring. It further has applications to microseismic monitoring, and to earthquake early warning. The latter is a particularly exciting application, because if the long-term memory abilities of RNN are fully utilized, wave propagation patterns made by foreshocks could be learned from dynamically and used by the RNN to make a high-confidence event detection of a large magnitude event with only one or two stations.

\section{Conclusions}
We have developed a new method for performing seismic phase association using deep learning. The method has been shown to achieve outstanding performance on an extremely active aftershock sequence in southern California, where many of the events are only seconds apart in origin time. It can be trained using only synthetic seismic phase arrival time data, and can therefore be applied even to networks that lack large amounts of labeled training data. The method does not use grids in any form to solve the association problem -- it instead learns to scan sequences of picks and identify patterns that resemble those of seismic wave propagation in the Earth.

\section*{Acknowledgements}
This research was supported by an artificial intelligence research grant from Amazon Web Services. The authors thank Pascal Audet and an anonymous reviewer for their helpful comments that improved the manuscript. The authors were also supported by the Gordon and Betty Moore Foundation, the Swiss National Science Foundation and the NSF Geoinformatics program. We have used waveforms and meta-data data from the Caltech/USGS Southern California Seismic Network (SCSN), doi: 10.7914/SN/CI; stored at the Southern California Earthquake Data Center, doi:10.7909/C3WD3xH1. We used TensorFlow \cite{tensorflow2015-whitepaper} and keras \cite{chollet2015keras} for all deep learning computations.

\bibliographystyle{unsrt}
\bibliography{phaselink_arxiv}

\end{document}